\documentclass[conference]{IEEEtran}
\IEEEoverridecommandlockouts

\usepackage{subcaption}
\usepackage{caption}
\usepackage[utf8]{inputenc}
\usepackage{amsmath}

\usepackage{booktabs}
\usepackage{multirow}

\usepackage[binary-units=true]{siunitx}

\usepackage[algoruled,noline,longend,linesnumbered]{algorithm2e}

\usepackage{pgfplots}
\usepackage{pgfplotstable}
\usepackage{siunitx}

\makeatletter
\usepackage{enumitem}
\setlist{topsep=0.2em, itemsep=0.3em}
\usepackage{amsthm}
\usepackage[small,compact]{titlesec}
\titlespacing{\section}{1pt}{1pt}{1pt}
\titlespacing{\subsection}{0.5pt}{0.5pt}{0.5pt}
\titlespacing{\subsubsection}{0.5pt}{0.5pt}{0.5pt}

\begin{document}

\graphicspath{{Fig/}}
\def\figname{Figure}
\def\algname{Algorithm}
\newcommand{\papertitle}{SteppingNet: A Stepping Neural Network with Incremental Accuracy Enhancement}

\title{\papertitle}
\author{
\IEEEauthorblockN{Wenhao Sun$^1$,
Grace Li Zhang$^2$,
Xunzhao Yin$^3$,
Cheng Zhuo$^3$,
Huaxi Gu$^4$,
Bing Li$^1$,
Ulf Schlichtmann$^1$}
\IEEEauthorblockA{$^1$Technical University of Munich (TUM), $^2$TU Darmstadt, $^3$Zhejiang University, $^4$Xidian University}
\IEEEauthorblockA{Email: \{wenhao.sun, b.li, ulf.schlichtmann\}@tum.de, grace.zhang@tu-darmstadt.de,}
\IEEEauthorblockA{\{xzyin1, czhuo\}@zju.edu.cn, hxgu@xidian.edu.cn}
}

\maketitle


\begin{abstract}

Deep neural networks (DNNs) have successfully been applied in many fields in
  the past decades. 
However, the increasing number of multiply-and-accumulate (MAC) operations in
  DNNs prevents their application  in resource-constrained and resource-varying
  platforms, e.g., mobile phones and autonomous vehicles.  In such platforms,
  neural networks need to provide acceptable results quickly and the accuracy
  of the results should be able to be enhanced dynamically
  according to the computational
  resources available in the computing system.  To address these
  challenges, we propose a design framework 
called SteppingNet. SteppingNet constructs a series of subnets whose accuracy
  is incrementally enhanced as more MAC operations become available.  Therefore, this design
  allows a trade-off between accuracy and latency.  In addition, the larger
  subnets in SteppingNet are built upon smaller subnets, so that the results of
  the latter can directly be reused in the former without recomputation.  This
  property allows SteppingNet to decide on-the-fly whether to enhance the
  inference accuracy by executing further MAC operations.
Experimental results demonstrate that SteppingNet provides an effective
  incremental accuracy improvement and its inference accuracy consistently
  outperforms
    the state-of-the-art work
    under the same limit of computational resources.

\end{abstract}

\section{Introduction} \label{sec:intro}

In recent years, deep neural networks (DNNs) have achieved remarkable
breakthroughs in many fields, e.g., image and speech recognition.  This
advance, however, is achieved at the cost of increasing number of
multiply-and-accumulate (MAC) operations.  For example, ResNet with 152 layers
\cite{Kaiming_CVPR_2016} requires 11.3G MAC operations to achieve its high
inference accuracy.  This tremendous computational cost poses 
challenges when DNNs are applied in resource-constrained and resource-varying
platforms, e.g., mobile phones and autonomous vehicles.  

The challenges are two-fold. First, these platforms
require a fast response time with a limited amount of computational resources.
For
example, in autonomous vehicles, it is crucial that potential emergencies are
recognized quickly to allow the vehicles to respond proactively.  However, 
the inference of neural networks in such vehicles may take longer than acceptable. 
E.g., according to \cite{Kocic_2019}, AlexNet takes 26ms on NVIDIA GTX 1070Ti. 
Proportionally, VGG-16 can take 780ms in inference, too large for autonomous driving \cite{Lin-2018}. 
Second, in
such platforms, computational resources vary dynamically due to the tasks
executed in parallel.
This requires that neural networks should be flexible in refining the inference
results with newly available resources instead of reexecuting all the MAC
operations from scratch.  For instance, the switch between normal mode and
power-saving mode of mobile phones leads to a change of available
computational resources \cite{Satish_2013},  so that neural networks
executed on such platforms should be able to adapt themselves with respect to
available resources dynamically.

To address the challenges described above,
\cite{Howard_2017,Howard_2018,Xiangyu_2017} propose efficient models that
provide a global hyperparameter, called width multiplier, to scale neural
networks for mobile applications, so that a trade-off between accuracy and
latency can be made.  However, these models require a large offline table to
store several models 
simultaneously.  In contrast, recent work \cite{Caihan_ICLR_2020,
Kim_CVPR_2018,Jiahui_ICLR_2019,Jiahui_ICCV_2019,Huang2018MultiScaleDN,Thanh_CVPR_2020}
trains a shared neural network
consisting of a series of subnets that have different numbers of weights and
thus different numbers of MAC operations. Since the weights are shared among
subnets, only one copy of the neural network needs to be stored.  During
inference, these subnets can 
be selected
according to the
current computational resources.


To implement a shared neural network, in \cite{Caihan_ICLR_2020} a once-for-all
network is trained as a whole and specialized subnets are generated
by selecting only a part of the once-for-all network according to resource
constraints of a hardware platform.  In \cite{Kim_CVPR_2018}, the NestedNet has
an $n$-in-$1$-type nested structure, which consists of $n$ subnets with
different sparsity ratios.  In addition, the slimmable network in
\cite{Jiahui_ICLR_2019,Jiahui_ICCV_2019} introduces a single neural network
that can be trained to operate at $N$ modes.  The subnets of different modes
provide a trade-off between latency and accuracy.  
In \cite{Huang2018MultiScaleDN}, a multi-scale neural network, each layer of which has a classifier, 
is designed to allow early-exits, so that subnets with different number of layers can be constructed. 
Furthermore, the any-width
network in \cite{Thanh_CVPR_2020} proposes a neural network with a single
training and subnets are constructed by selecting different widths of neuron
connections, so that a fine-grained control over accuracy and latency during
inference can be achieved.

The previous work
\cite{Caihan_ICLR_2020,Kim_CVPR_2018,Jiahui_ICLR_2019,Jiahui_ICCV_2019} can
achieve a trade-off between accuracy and latency, but they are designed to
select a subnet according to the current available computational 
resources statically. If more
resources become available after the execution of the selected subnet has been
started, these resources cannot be taken advantage of to enhance the inference
accuracy by switching to a larger subnet without discarding the current
intermediate results.  
Although the multi-scale network in \cite{Huang2018MultiScaleDN} and 
the any-width network in \cite{Thanh_CVPR_2020}
allow a dynamic adjustment of the subnet by executing more predetermined MAC
operations on newly available resources, 
the structures of the subnets are
severely restricted to allow the free expansion of subnets. Accordingly, the
inference accuracy of the subnets is negatively affected.

To provide both flexibility and capability of computational reuse in inference,
we propose a design framework, called SteppingNet, for neural networks executed
on resource-constrained and resource-varying platforms.  The contributions of
this work are summarized as follows.
\begin{itemize} 
  \item 
    SteppingNet constructs a series of subnets, whose structures are 
    adapted 
    according to
    the allowed numbers of MAC operations. The accuracy of these subnets
    is incrementally enhanced, so that they can provide a good trade-off
    between accuracy and latency in resource-constrained and resource-varying
    platforms. 
  \item
    SteppingNet maximally exploits computational reuse among subnets.  The
    intermediate results of a subnet can directly be reused in subsequent
    larger subnets to improve inference accuracy
    in case computational resources become available
    dynamically.
  \item 
    With the MAC-constrained structures and the incremental nature of
    the subnets, SteppingNet is very suitable for important scenarios 
    where a preliminary decision should be made 
    early and refined further with more computational resources or
    execution time.
  \item 
    Experimental results demonstrate that SteppingNet provides an effective
    incremental accuracy improvement with respect to invested computational
    resources.  Compared with state-of-the-art work, SteppingNet can
    achieve a consistently better accuracy under the same limit of
    computational resources.
\end{itemize}

The rest of this paper is organized as follows. In
Section~\ref{sec:motivation}, the background and motivation of this work are
explained.  The proposed framework to determine a series of subnets with
incremental accuracy enhancement is explained in Section~\ref{sec:framework}.
Experimental results are reported in Section~\ref{sec:results} and conclusions
are drawn in Section~\ref{sec:conclusion}.

\begin{figure}
  \includegraphics{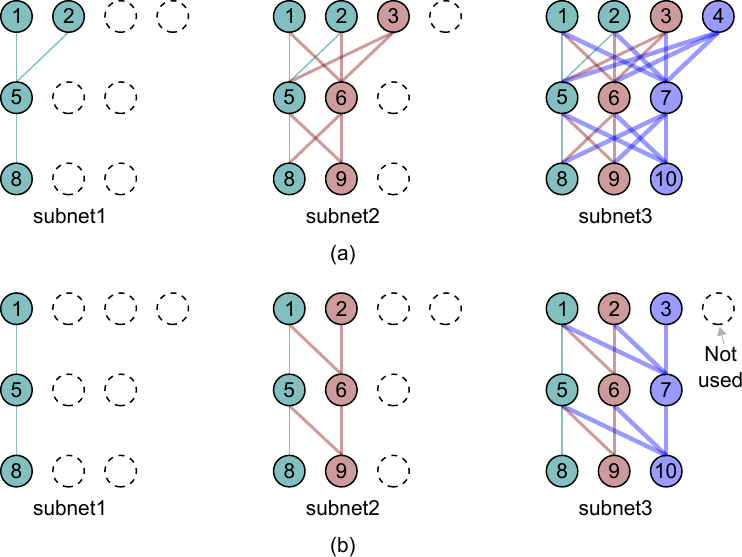}
  \caption{Structures of the slimmable network
  \cite{Jiahui_ICLR_2019} and the any-width network \cite{Thanh_CVPR_2020}.  
  (a) Three subnets in the slimmable network. (b) Three subnets in
  the any-width network. }
  \label{fig:example}
\end{figure}

\section{Background and Motivation}\label{sec:motivation}

When neural networks are applied for inference in resource-constrained and resource-varying
platforms, they should be flexible and adaptive to the varying
computational resources.  To achieve this goal, the slimmable network in
\cite{Jiahui_ICLR_2019} proposes to train several subnets, which
share the same set of weights. When deployed into a computing system, one of
the subnets is selected according to the available computational resources to
provide a trade-off between accuracy and computational cost.
\figname~\ref{fig:example}(a) illustrates the concept of the slimmable network,
where three subnets with different numbers of weights 
are designed
in advance. A node in this example represents either a neuron or a filter,
depending whether it is a fully-connected layer or a convolutional layer.  For
convenience, we will refer to such a node as a neuron henceforth. In the slimmable
network, the inputs to a neuron in different subnets can be different, e.g.,
neuron 5 in \figname~\ref{fig:example}(a). Accordingly, different batch 
normalization layers need to be stored for the subnets during the inference phase.

In the slimmable network, larger subnets may invalidate the computational results
at neurons in a smaller subnet. Consequently, intermediate results of the
smaller subnet cannot directly be reused in larger subnets. 
For
example, in \figname~\ref{fig:example}(a), the synapse from neuron 3 to neuron
5 in subnet2 requires the recomputation of neuron 5 in subnet2 during inference. 
This
problem is overcome in the any-width network \cite{Thanh_CVPR_2020}, in which no
synapse connects a neuron that is only in a larger subnet to another neuron in
a smaller subnet, as illustrated in \figname~\ref{fig:example}(b). 
Accordingly, 
the any-width network does not require extra batch
normalization layers for subnets. More importantly, the any-width network allows
dynamic expansion of subnets. Once extra computational resources are available,
the any-width network can always switch to the next larger subnet to enhance
the inference accuracy by executing more MAC operations in the expanded network. 
Similarly, when the computational resources reduce dynamically, 
the smaller subnet can also reuse the intermediate results of the previous larger subnet.

\begin{figure}
  \includegraphics{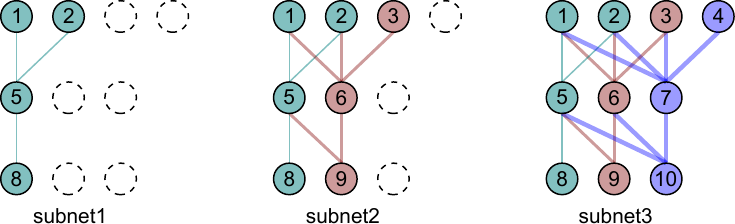}
  \caption{Structures of the proposed SteppingNet with three subnets.}
  \label{fig:proposed}
\end{figure}

Despite its advantages in dynamic subnet expansion, the any-width network still
suffers from limitations. First, the structures of the subnets are manually
determined. These structures must follow the regular pattern shown in
\figname~\ref{fig:example}(b).  This strict structural pattern, however, may
impair the inference accuracy of subnets, because further potential network
structures of subnets are not explored.  For example, a subnet can have an
irregular structure, such as subnet$1$ in \figname~\ref{fig:proposed}.  When
this subnet is expanded into subnet$2$, it only needs to be guaranteed that
newly expanded neurons do not have synapses starting from them and entering the
neurons in subnet$1$ to maintain the capability of dynamic subnet expansion and reduction.
Second, the regular structures in the any-width network may not use up all the
neurons in the neural network as shown in \figname~\ref{fig:example}(b), where
neuron 4 cannot be included into a subnet if the regular structural pattern is
strictly followed. Third, since the any-width network constructs the subnets according to
structural rules, the computational cost in each expansion of the
subnets, i.e., the extra MAC operations, is not directly controlled.
Consequently, when used in computing systems with dynamically varying resources, 
the expansion of subnets in the any-width network may not work due to mismatch of the
required and the dynamically available computational resources.

\section{Construction and Training of SteppingNet}\label{sec:framework}

\begin{figure}
  \includegraphics{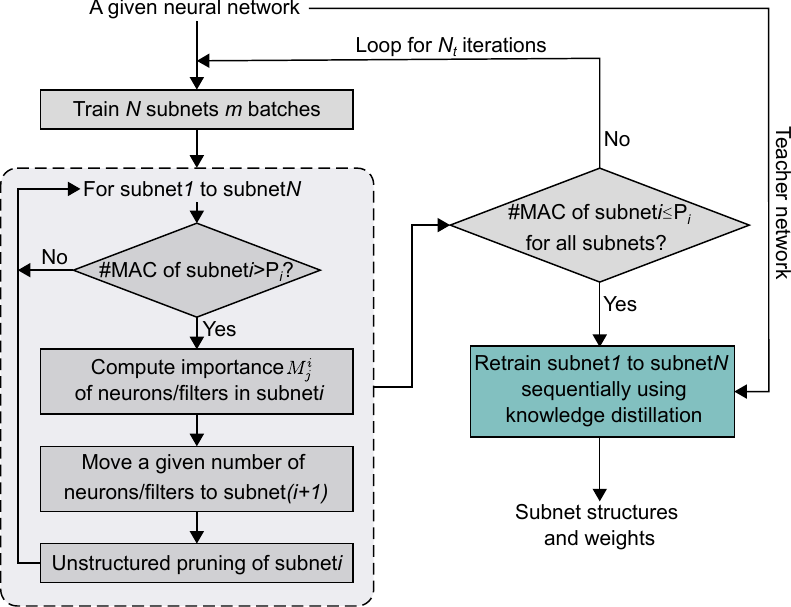}
  \caption{Work flow of SteppingNet. Except training multiple subnets with
  knowledge distillation in the last step, the other steps belong to subnet
  construction.}
  \label{fig:flow}
\end{figure}

To implement a series of efficient subnets whose weights are shared and whose
inference accuracy is enhanced incrementally when more computational
resources become available, we use the work flow shown in \figname~\ref{fig:flow}. The
construction process determines the structures of the subnets by moving neurons
gradually between subnets. In the last step, the subnets are retrained with
knowledge distillation to enhance their inference accuracy. The construction
and retraining of subnets are described in detail in Section~\ref{sec:search}
and  Section~\ref{sec:training}, respectively.

\subsection{Constructing subnets of SteppingNet by neuron assignment}
\label{sec:search}

The task of subnet construction is to determine the structures of the
subnets.  A smaller subnet should be contained in a larger subnet so that its
results can contribute to the computation results of the latter. In addition,
the extra neurons in the  larger subnet should not have synapses to the neurons
in the smaller subnet; otherwise, the neurons in the smaller subnet need to be
reevaluated, thus losing the incremental property of subnets.

\begin{figure}
  \includegraphics{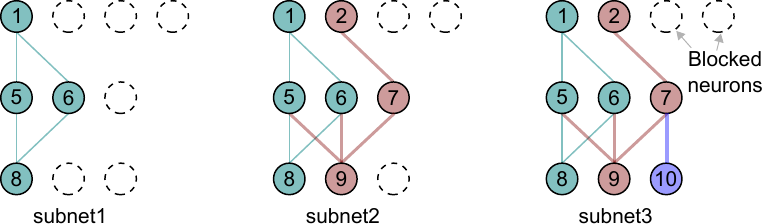}
  \caption{Inappropriate subnet construction with blocked neurons.}
  \label{fig:invalid_example}
\end{figure}

A straightforward idea of subnet construction is selecting weights 
according to their importance for each subnet. 
However, this method does not consider the incremental property of subnets, 
so that it 
can unfortunately block some neurons and 
lead to a suboptimal result.  
For
example, in \figname~\ref{fig:invalid_example}, 
weights that are important for subnet1 are selected from the original network. 
Subnet$2$ is constructed by selecting its important weights while 
guaranteeing that 
subnet$2$ does not have synapses to the neurons in subnet$1$. 
After the construction of subnet$1$ and subnet$2$, all the three neurons in the
second layer are already occupied by the two subnets.  Therefore, it is
not possible to include the remaining two neurons in the first layer into
subnet$3$ without invalidating the computation results of subnet$1$ and
subnet$2$. This limitation wastes the potential of neurons in subnets and thus
compromises the inference accuracy of subnet$3$.

To avoid the problem above, we will evaluate the importance of the neurons 
with respect to all subnets 
and move them across subnets to gradually build up
the structures of subnets while guaranteeing the incremental property.  
In the following, the process of moving neurons to construct subnets is
described Section~\ref{sec:construct_process} and the importance evaluation of
neurons is explained in Section~\ref{sec:importance_evaluation}.

\subsubsection{Structural construction of subnets}\label{sec:construct_process}

In SteppingNet, subnets are constructed from a given original neural network.
Each subnet contains a part of the neurons and synapses of the original neural
network, and a smaller subnet is completely contained in a larger subnet.  The
inference accuracy of the original neural network is an upper bound of the
inference accuracy of the subnets.  

In the construction process, the smallest subnet is first initialized with the
original neural network.  The neurons are gradually moved away from this subnet
to fill larger subnets using the work flow in \figname~\ref{fig:flow}.  In this
flow, the subnets are first trained for $m$ batches and their numbers
of MAC operations are evaluated sequentially
afterwards. 
If the number of MAC operations of subnet$i$ is larger than a
predefined threshold $P_{i}$, some neurons in subnet$i$ are moved into
subnet$(i+1)$ according to their importance to the subnets. The flow 
ends when the number of MAC operations of each subnet satisfies the requirement. 
During this process, the extra neurons in a larger subnet are not allowed 
to have synapses to the neurons in a smaller subnet, 
so that the capability of dynamic subnet 
expansion and reduction is maintained.

\begin{figure}
  \includegraphics{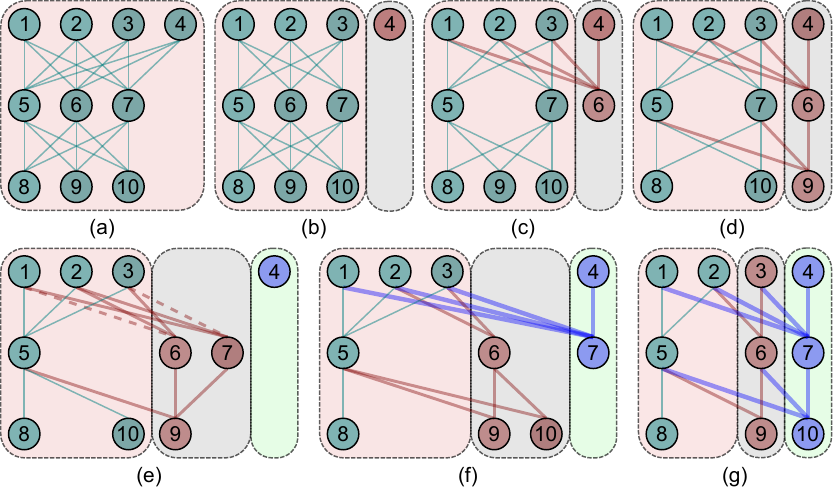}
  \caption{Subnet construction in SteppingNet. The leftmost block with neurons in green belongs to
  subnet$1$, the two blocks on the left with neurons in green and red (from (b) onwards) belong to subnet$2$, and the three
  blocks together with neurons in green, red and purple (from (e) onwards) form subnet$3$.  
  At the beginning, subnet$1$ is initialized
  with the original neural network and the other subnets are empty in (a).
  }
  \label{fig:construction_process}
\end{figure}

An example of the construction process is illustrated in
\figname~\ref{fig:construction_process}, where subnet$1$ is initialized 
using the original neural network as
shown in \figname~\ref{fig:construction_process}(a).  Assume that the allowed
MAC operations in the three subnets to be constructed in
\figname~\ref{fig:construction_process} are 3, 7, 14, respectively.  The number
of MAC operations in subnet$1$ in \figname~\ref{fig:construction_process}(a)
thus needs to be reduced.  Accordingly, neuron 4 is moved to subnet$2$ in
\figname~\ref{fig:construction_process}(b).  This process is repeated for 
subnet$1$ and neurons gradually flow into subnet$2$. When the difference in the numbers of MAC 
operations of subnet$2$ and subnet$1$ is larger than $7-3=4$, the neurons start to flow from subnet$2$ to 
subnet$3$; Otherwise 
subnet$2$ cannot maintain a sufficient number of neurons, so that 
the number of 
MAC operations in subnet$2$ might be much smaller than the allowed number at the end of construction process. 
In \figname~\ref{fig:construction_process}(d), 
the MAC difference of subnet$2$ and subnet$1$ is more than 4, 
so that 
a neuron
is moved from subnet$2$ to subnet$3$ in
\figname~\ref{fig:construction_process}(e), while 
neuron movement from subnet$1$ to subnet$2$ is ongoing simultaneously.
After the iterations in
\figname~\ref{fig:flow} are finished, the structures of the subnets are 
determined, as illustrated in \figname~\ref{fig:construction_process}(g). 

In moving a neuron from subnet$i$ to subnet$(i+1)$, all the synapses from this
neuron to the neurons in subnet$i$ are removed to avoid the reevaluation of the
neurons in subnet$i$. For example, in
\figname~\ref{fig:construction_process}(b), neuron 4 loses all the connections
to the neurons in the second layer of the neural network.  However, when
further neurons are moved into subnet$(i+1)$, the synapses between the neurons
are reestablished to maintain the inference accuracy, e.g., the synapse between
neuron 4 and neuron 6 in  \figname~\ref{fig:construction_process}(c). 

After neurons in the subnets are updated in an iteration, we also apply
pruning \cite{Song_ICLR_2016} to remove the weights and filters that are
unimportant to the inference accuracy of the corresponding subnet, such as the
synapse between neuron 1 and neuron 6 and  the synapse between neuron 3 and
neuron 7 in \figname~\ref{fig:construction_process}(e).  Consequently, weights
and the corresponding MAC operations remaining in a subnet essentially
contribute to its inference accuracy. 
Since weights pruned in subnet$i$ may be important to larger subnets,
we do not remove these weights permanently during pruning but allow them to
update in the following training iterations, so that the importance of neurons
to larger subnets can be evaluated correctly, as explained in
Section~\ref{sec:importance_evaluation}.  When a neuron with pruned weights is
moved to another subnet, the corresponding synapses are revived, because these
synapses may be essential to the new subnet, such as the synapse between
neuron 3 and neuron 7 in \figname~\ref{fig:construction_process}(f).  

In the example in \figname~\ref{fig:construction_process}, only one neuron from
a subnet is moved to the next subnet in an iteration.  
Since the number of
neurons/filters in a deep neural network can be large, in implementing
SteppingNet we move multiple neurons simultaneously from a subnet to another.
We first determine the number of MAC operations required to be moved from a
subnet to the
subsequent subnet according to the allowed number of MAC operations.  
Since subnet$1$ needs to move the largest number of neurons to other subnets,
we use it to calculate an upper bound of the number of
neurons moved between subnets in an iteration.
Assume
that the number of MAC operations allowed in subnet$1$ is $P_1$, the total
number of MAC operations of the original neural network is $P_t$, and the total
number of iterations allowed in the work flow in \figname~\ref{fig:flow} is
$N_t$. The number of MAC operations that are moved from a subnet to the
subsequent larger
subnet in an iteration is then defined as $(P_t-P_1)/N_t$ to guarantee
that the final numbers of remaining MAC operations in the subnets
comply with the requirements.
In implementing SteppingNet, the neurons are evaluated 
and 
a set of neurons whose importance
values are low and whose number of MAC operations just exceeds $(P_t-P_1)/N_t$  
are selected from subnet$i$ and moved to subnet$(i+1)$.


\subsubsection{Importance evaluation for neuron reallocation}
\label{sec:importance_evaluation}

In SteppingNet, if a neuron appears in subnet$i$, it also appears in all
the larger subnets. The importance of this neuron
with respect to different subnets
may be different.
Accordingly, we use a parameter $r_j^{i}$ to indicate the importance
of the $j$th neuron with respect to subnet$i$. 
With $r_j^{i}$, we then modify the computation 
at the $j$th neuron in the neural network as 
\begin{align}\label{eq:importance}
  d_j=\varphi(r_j^{i}*\sum_{k=0}^{n_j} d_{j,k}*w_{j,k}+b_j)
\end{align}
where $d_j$ is the output of this neuron.
$d_{j,k}$ and $w_{j,k}$ are the input and the weight of the $k$th synapse
to this neuron, respectively.    $n_j$ is
the total number of incoming synapses to this neuron.  $b_j$ is the bias.
$\varphi$ is the activation function.  For CNNs, $r_j^{i}$ is assigned to the
$j$th filter of the $i$th subnet to indicate the importance of this filter. The
operation of the filter is the corresponding convolution instead of the MAC
operation in (\ref{eq:importance}).

During forward propagation, $r_j^{i}$ is set to 1 to guarantee the correct
function of this neuron in inference. Since the subnets should be trained
to enhance inference accuracy,  
we maintain a cost function $L_i$ for each subnet.  At
backward propagation, we calculate the partial derivative of the cost function
$L_i$ to $r^i_j$ 
as
\begin{equation}
\frac{\partial L_i}{\partial r_j^{i}}=\frac{\partial L_i}{\partial \varphi}
\frac{\partial \varphi}{\partial (r^i_j*\sum_{k=0}^{n_j}
  d_{j,k}*w_{j,k}+b_j)}\times \sum_{k=0}^{n_j} d_{j,k}*w_{j,k}.
\end{equation}
In backward propagation, $\frac{\partial L_i}{\partial r_j^{i}}$ is a floating-point
number, which
we use 
to make the binary decision whether a
neuron should be moved to the next larger subnet.

In selecting a neuron to move from subnet$i$ to subnet$(i+1)$, 
$\frac{\partial L_i}{\partial r_j^{i}}$
does not provide a sufficiently good indication,
since a neuron in subnet$i$ is also contained in all the
subnets larger than subnet$i$. 
Accordingly, 
we tend to keep the neurons that are also important to all the larger subnets,
and define the selection
criterion for the $j$th neuron in the $i$th subnet as
\begin{equation}\label{eq:mij}
  M^i_j=\sum_{k=i}^N \alpha_k\left|\frac{\partial L_k}{\partial r_j^k}\right|
\end{equation}
where $\alpha_k$ is a constant defining the contribution ratio of 
$\frac{\partial L_k}{\partial r_j^{k}}$
of a neuron with respect to subnet$k$. $N$ is the number of
subnets. In an iteration, after
$M^i_j$ are updated, the neurons with the smallest $M^i_j$ are moved to the
next subnet. 
\begin{table*}
\centering
{
\caption{Results of SteppingNet}
\vspace*{-8pt}
\label{tb:results}
\begin{tabular}{cc c c c  ccc c ccc c ccc} \hline
\multicolumn{2}{c}{Test cases} &
\multicolumn{1}{c}{} &
\multicolumn{1}{c}{Orig. Net} &
\multicolumn{1}{c}{} &
\multicolumn{2}{c}{Subnet1} &
\multicolumn{1}{c}{} &
\multicolumn{2}{c}{Subnet2} &
\multicolumn{1}{c}{} &
\multicolumn{2}{c}{Subnet3} &
\multicolumn{1}{c}{} &
\multicolumn{2}{c}{Subnet4} \\
\cline {1-2} \cline{3-4} \cline{6-7} \cline{9-10} \cline{12-13} \cline{15-16}  
\multicolumn{1}{c}{Network} &
\multicolumn{1}{c}{Dataset} &
\multicolumn{1}{c}{} &
\multicolumn{1}{c}{Acc.} &
\multicolumn{1}{c}{} &
\multicolumn{1}{c}{$A_1$} &
\multicolumn{1}{c}{$M_{1}/M_t$} &
\multicolumn{1}{c}{} &
\multicolumn{1}{c}{$A_2$} &
\multicolumn{1}{c}{$M_{2}/M_t$} &
\multicolumn{1}{c}{} &
\multicolumn{1}{c}{$A_3$} &
\multicolumn{1}{c}{$M_{3}/M_t$} &
\multicolumn{1}{c}{} &
\multicolumn{1}{c}{$A_4$} &
\multicolumn{1}{c}{$M_{4}/M_t$}\\
\hline

LeNet-3C1L    &Cifar10    &&83.36\%  &&68.5\% &9.65\%     &&77.38\% &29.55\%   &&79.81\% &48.62\% &&80.4\% &78.52\%    \\
LeNet-5       &Cifar10    &&74.96\%   &&51.8\% &13.64\%   &&59.56\% &26.54\%   &&68.64\% &55.07\% &&72.03\% &82.74\%   \\
VGG-16       &Cifar100    &&70.32\%   &&63.26\% &15.97\%  &&68.19\% &32.54\%   &&68.19\% &47.39\% &&68.14\% &67.78\%   \\

\hline
\end{tabular}
}
\end{table*}


Because moving neurons between subnets changes the structures as
well as the cost functions of the subsets, we train the subnets with $m$
batches 
before evaluating the
neurons using (\ref{eq:mij}), as shown in \figname~\ref{fig:flow}.
In this process, the training of a larger subnet also updates the weights in
smaller subnets, whose values have been determined by directly training the smaller
subnets themselves in the same iteration.  
Consequently, the accuracy of smaller subnets may
degrade after a larger subnet is trained.  To reduce the effect of updating
weights in smaller subnets when training a larger subnet, we decrease the
learning rate of weights in a smaller subnet by the ratio $\beta^{(j-i)}$,
where $\beta$ is a constant between 0 and 1 and
$j$ and $i$ are the indexes of the larger and the smaller subnets,
respectively.  
$(j-i)$ is used as the exponent of $\beta$ so that the smaller the subnets are,
the more their learning rates are decreased.
With this
reduction of learning rates, smaller subnets obtain more stability to maintain
their inference accuracy when larger subnets are trained.

\subsection{Retraining subnets with knowledge distillation}
\label{sec:training}

After subnets are constructed with the method in Section~\ref{sec:search}, we
retrain them to improve their inference accuracy with knowledge distillation
\cite{Ying_CVPR2018}.  In this retraining, the teacher network is the original
neural network from which subnets are constructed.  This neural network has a
high accuracy compared with the subnets, which are the student networks, so
that it is used to guide the subnets during retraining. 

When retraining a subnet, we modify its cost function as follows 
\begin{align}
L_i^\prime &= \gamma\times L_i+(1-\gamma)
  \sum_{k=1}^{n_c} Y_k log(\frac{Y_k^{pre}}{Y_k}) \label{eq:KL}
\end{align}
where $L_i$ is the cross entropy of subnet$i$.  $ \sum_{k=1}^{n_c} Y_k
log(\frac{Y_k^{pre}}{Y_k})$ is the Kullback-Leibler divergence \cite{Joyce2011}
of subnet$i$ to the pretrained original neural network, where $Y^{pre}_k$ and
$Y_k$ are the $k$th outputs of the original neural network and subnet$i$,
respectively. $n_c$ is the number of output classes.  $\gamma$ is a
constant between 0 and 1 to adjust the priority of Kullback-Leibler divergence in
(\ref{eq:KL}).
The smaller the difference between $Y_k^{pre}$ and $Y_k$ is, the more
similar results the subnets generate compared with the original neural network.

In the retraining phase, we train the subnets in an ascending order in each
epoch using the modified cost function in (\ref{eq:KL}). During this training, we
also reduce the learning rates of subnets 
as described in
Section~\ref{sec:importance_evaluation} to avoid drastic weight change in
smaller subnets. With this multi-subnet knowledge distillation, the inference
accuracy of all these subnets can be enhanced and balanced as a whole.

\section{Experimental Results}\label{sec:results}
To evaluate the effectiveness of SteppingNet,
three neural networks, LeNet-3C1L, 
LeNet-5 
and VGG-16 were 
applied onto two datasets, 
Cifar10 and Cifar100, respectively.  
as shown in the first two columns of Table \ref{tb:results}. 
The construction of subnets and their retraining  
were implemented using PyTorch and tested on Nvidia Quadro RTX 6000 GPUs. 



To demonstrate that SteppingNet provides an incremental accuracy improvement
with respect to 
computational resources, four subnets were constructed and retrained with the
framework described in Section~\ref{sec:framework}.  During this construction,
the connections from new neurons in a subnet to the neurons in the  previous
smaller subnets were prohibited to enable computational reuse as described in Section III. 
To provide the construction process more flexibility, 
we expanded the number of neurons/filters of each layer in the original
network as in \cite{Thanh_CVPR_2020}
and initialized the first subnet in the construction process with this expanded
network.  For LeNet-3C1L, LeNet-5, and VGG-16, the corresponding expansion
ratios were set to 1.8, 2.0, 1.8, respectively. For example, 
in the expanded LeNet-3C1L, 
the number of neurons/filters is 1.8 times of that of the original network. 
\begin{figure}
    \centering
    \begin{subfigure}[t]{0.35\textwidth}
        \centering
        \includegraphics[width=0.9\textwidth]{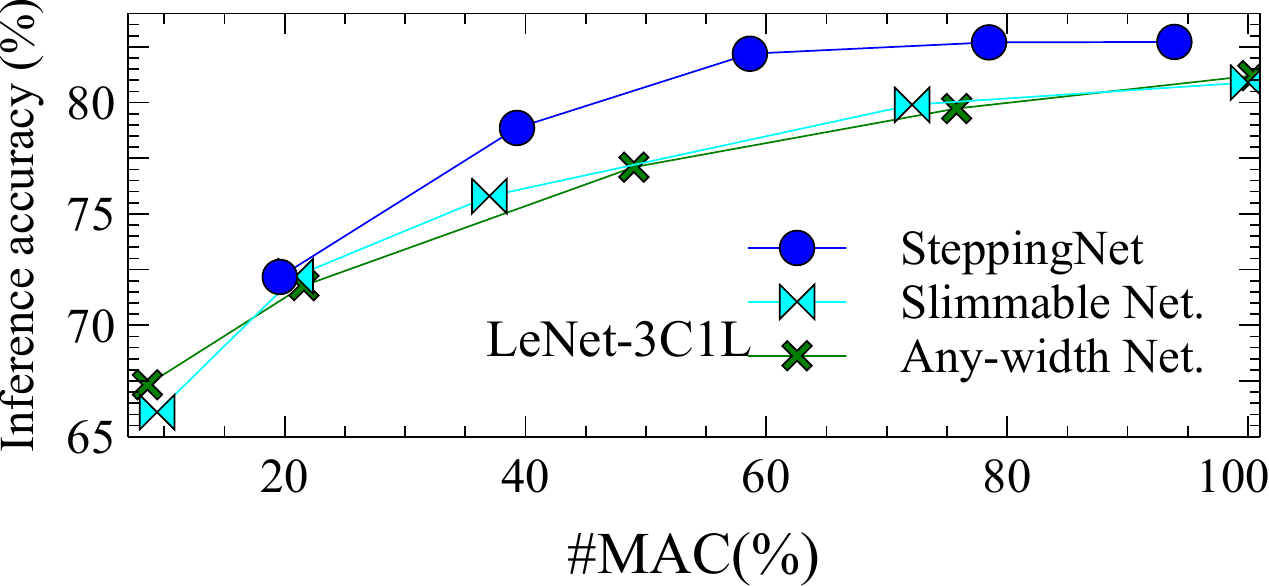}
    \end{subfigure}%

    \begin{subfigure}[t]{0.35\textwidth}
        \centering
        \includegraphics[width=0.9\textwidth]{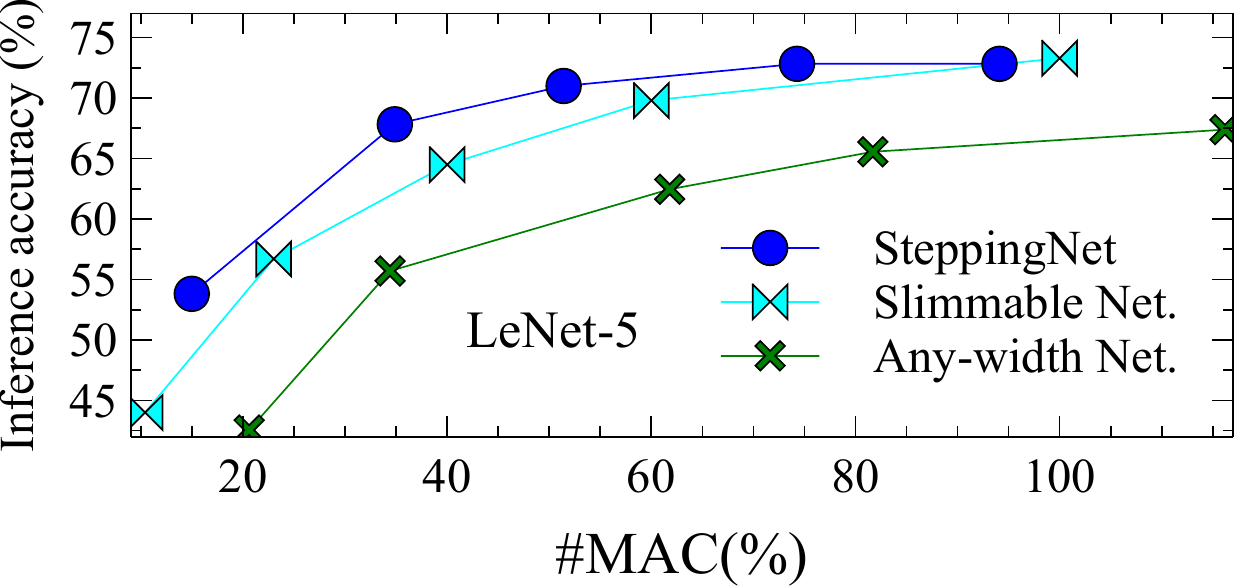}
    \end{subfigure}

    \begin{subfigure}[t]{0.35\textwidth}
        \centering
        \includegraphics[width=0.9\textwidth]{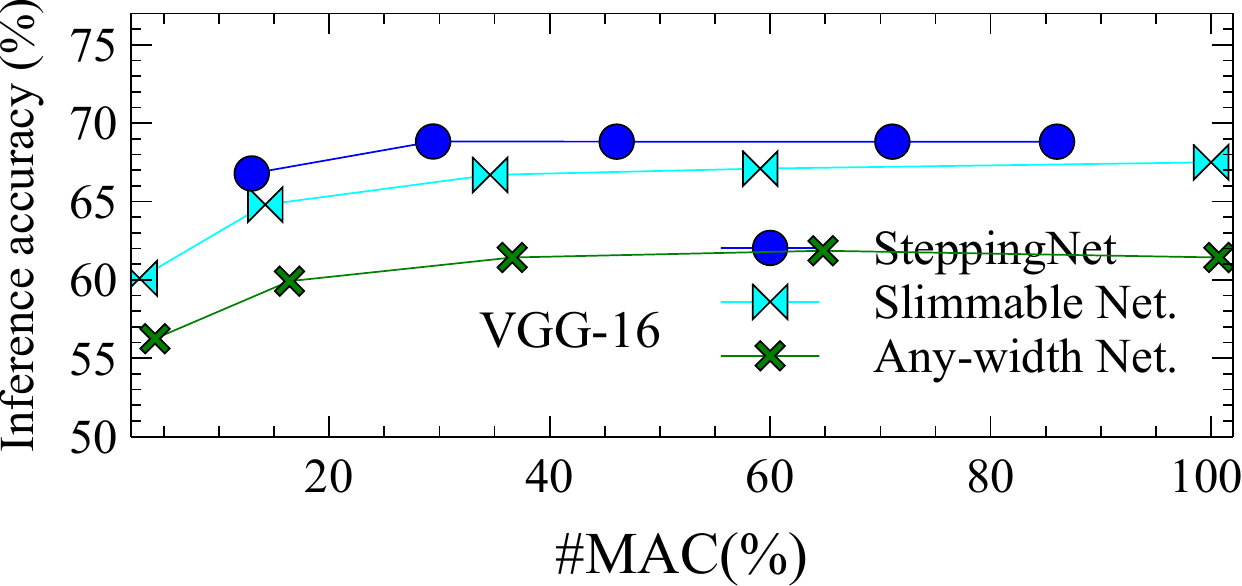}
    \end{subfigure}
    \caption{Comparison with the any-width network and the slimmable network.}
    \label{fig:comparison}
\end{figure}
In the flow in \figname~\ref{fig:flow}, we set the number of
training batches at the beginning of each iteration to 250, 250, and 100 for
LeNet-3C1L, LeNet-5, and VGG-16, respectively. The total number of allowed
iterations $N_t$ was set to 300.  The weight threshold in the unstructured
pruning was set to $1 \times 10^{-5}$.  The coefficients $\alpha_k$ in
(\ref{eq:mij}) were increased to 1.5 times from $\alpha_1=1$ for each larger subnet 
to emphasize the importance of the neurons to these larger subnets, so that
the neurons remaining in the current subnet also make good contribution to the
inference accuracy of the larger subnets.  $\beta$ in weight update suppression
in Section~\ref{sec:importance_evaluation} was set to 0.9.
$\gamma$ in (\ref{eq:KL}) was set to 0.4
to balance the cross entropy and the effect of knowledge distillation in
retraining.

During subnet construction and retraining, the inference accuracy of the
largest subnet could not be improved further after a certain number of
MAC operations was reached. 
For example, in LeNet-3C1L and LeNet-5, after executing around 85\% MAC
operations of the original network, the inference accuracy of the largest
subnet reached more than 95\% of the inference accuracy of the original network
and did not increase any further.  For VGG-16, this threshold was around 70\%.
Accordingly, we set these numbers of MAC operations as the resources allowed in
the largest subnets and decreased from these numbers to set the
resources allowed in smaller subnets, so that the subnets can produce inference accuracy
at different levels of MAC operations. 
For the three neural networks, LeNet-3C1L, LeNet-5 and
VGG-16, the allowed MAC operations in the four subnets were then set to
10\%/30\%/50\%/85\%, 15\%/30\%/60\%/85\%, 20\%/40\%/50\%/70\% of the
original neural networks, respectively.

Table \ref{tb:results} shows the inference accuracy of the test cases.  The
third column shows the inference accuracy of the original neural networks.  The
columns $A_1$, $A_2$, $A_3$, and $A_4$ show the inference accuracy of the subnets.
The columns $M_{1}/M_t$, $M_{2}/M_t$, $M_{3}/M_t$ and $M_{4}/M_t$ show the
percentages of MAC operations in the subnets 
with respect to the number of MAC
operations $M_t$ of the original neural network.  
According to this table, it can be observed that the inference accuracy of
subnets was improved by more MAC operations.  In addition, the incremental accuracy
enhancement was not necessarily linear with respect to the number of MAC
operations.  For example, with even about 10\% of the total number of MAC
operations, the inference accuracy of LeNet-3C1L can already reach 68.5\%, 
which is important for some scenarios, e.g., autonomous driving, to make
a preliminary decision.  The inference accuracy of the largest subnets was
already close to that of the original neural networks, and the difference was 
to provide the property of incremental
enhancement.

To demonstrate the performance of SteppingNet compared with the any-width network
\cite{Thanh_CVPR_2020} and the slimmable network\cite{Jiahui_ICLR_2019}, 
we executed the any-width network and the slimmable network 
on the three networks in Table \ref{tb:results} to obtain
the inference accuracy of five subnets under various numbers of MAC operations.  
The accuracy comparison is illustrated in \figname~\ref{fig:comparison}. This comparison 
demonstrates that SteppingNet outperforms the any-width network and the slimmable network 
in inference
accuracy under the same numbers of MAC operations, due to the fact that
SteppingNet enables more flexible subnet structures than the any-width network and the slimmable network.


\begin{figure}
  \includegraphics[scale=0.35]{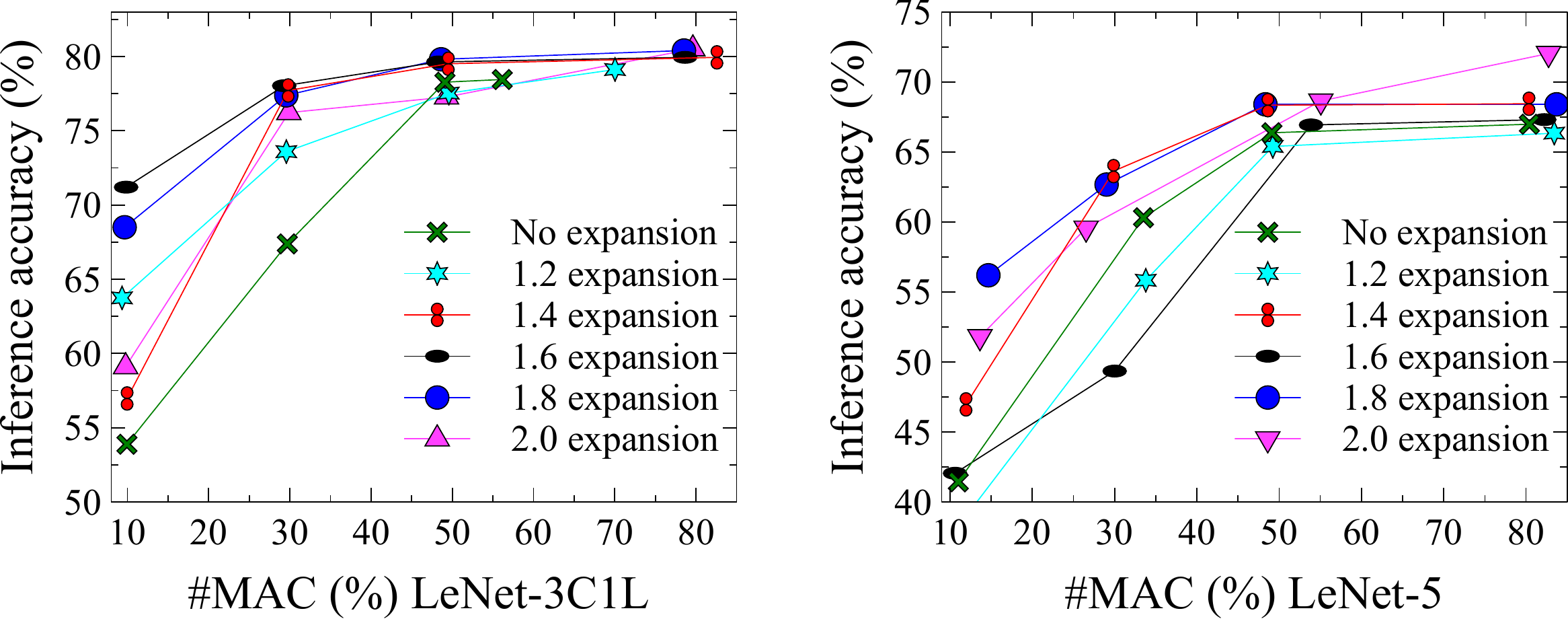}
  \caption{Accuracy comparison with different expansion ratios.}
  \label{fig:com_expansion}
\end{figure}

Before subnet construction, we expanded the number of neurons/filters
in the original network with a given ratio to allow more flexible subnet structures to
be identified.  To demonstrate how this ratio affects the inference accuracy,
we changed these ratios and tested the inference accuracy of the constructed
subnets, as shown in \figname~\ref{fig:com_expansion}, where the ratio of MAC
operations is 
with respect to the number of the MAC operations of the original neural network
without expansion.
According to this figure, it can be seen that different expansion ratios do
affect the accuracy of the subnets due to more available subnet structures.
The ratios that provide the best overall accuracy were selected to generate the
results in Table~\ref{tb:results}.

\begin{figure}
  \vskip 1.2pt
  \includegraphics[scale=0.35]{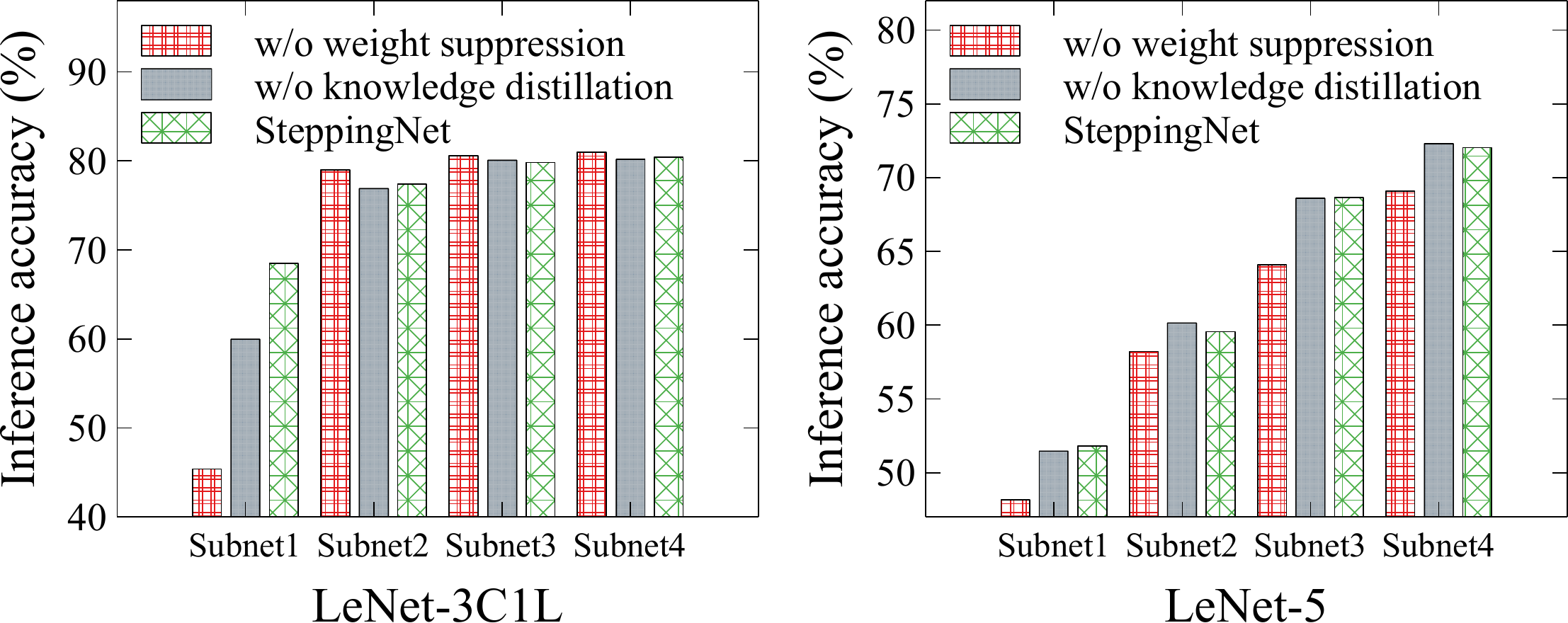}
  \caption{Accuracy comparison with and without suppression of weight update 
  and knowledge distillation.}
  \label{fig:com_KD}
\end{figure}

During the construction of subnets, we suppressed the update of weights in
smaller subnets to avoid accuracy loss 
when larger subnets were trained, as described in Section~\ref{sec:importance_evaluation}.
After subnet construction, we adopted knowledge distillation to retrain the
subnets.  To demonstrate the effectiveness of these techniques, we compared the
inference accuracy with these techniques disabled individually.  The results
are shown in \figname~\ref{fig:com_KD}. According to this figure, both weight update
suppression and knowledge distillation contribute to the inference accuracy.
When these two techniques are combined, inference accuracy of many subnets,
especially the smaller ones, can be enhanced. For larger subnets,
these techniques may interfere with each other and lead to slight accuracy
fluctuation, but the overall accuracy still stays relatively stable.

\section{Conclusion}
\label{sec:conclusion}

In this paper, we have proposed a design scheme, called SteppingNet, for neural
networks executed on resource-constrained and resource-varying platforms.
SteppingNet constructs a series of subnets with different numbers of MAC
operations and the intermediate results
of smaller subnets in SteppingNet can be reused directly in subsequent larger
subnets. 
Experimental results demonstrated that SteppingNet outperforms state-of-the-art
work in inference accuracy under the same limit of computational resources.

\let\oldbibliography\thebibliography
\renewcommand{\thebibliography}[1]{%
\oldbibliography{#1}%
\fontsize{6.0pt}{6.0}\selectfont
\setlength{\itemsep}{0.1pt}%
}

\bibliographystyle{IEEEtran}
\bibliography{IEEEabrv,CONFabrv,bibfile}

\end{document}